\documentclass[10pt,twocolumn,letterpaper]{article}

\usepackage{iccv}
\usepackage{times}
\usepackage{epsfig}
\usepackage{graphicx}
\usepackage{amsmath}
\usepackage{amssymb}

\usepackage[page]{appendix} 

\usepackage{amsmath}
\usepackage{amssymb}

\usepackage{algorithm}
\usepackage{algpseudocode}

\usepackage{epsfig}
\usepackage{graphicx}
\usepackage{subcaption}

\usepackage{booktabs}
\usepackage{pifont}
\usepackage[table,x11names]{xcolor}

\usepackage{multicol}
\usepackage{multirow}

\newcommand{\comment}[1]{}
\newcommand{\tsdagger}{{\textsuperscript{\textdagger}}}

\newcommand{\mbf}[1]{\mathbf{#1}}

\newcommand\blfootnote[1]{%
  \begingroup
  \renewcommand\thefootnote{}\footnote{#1}%
  \addtocounter{footnote}{-1}%
  \endgroup
}

\definecolor{LightCyan}{rgb}{0.88,1,1}

\usepackage{setspace}
\usepackage[pagebackref=true,breaklinks=true,letterpaper=true,colorlinks,bookmarks=false]{hyperref}
\iccvfinalcopy 

\def\iccvPaperID{6249} 

\begin{document}

\title{Segmenter: Transformer for Semantic Segmentation}

\author{Robin Strudel\thanks{Equal contribution.}\\
  Inria\begin{NoHyper}\thanks{Inria, \'Ecole normale supérieure, CNRS, PSL Research University, 75005 Paris, France.}\end{NoHyper}\\
  \and
  Ricardo Garcia\footnotemark[1]\\
  Inria\footnotemark[2]\\
  \and
  Ivan Laptev\\
  Inria\footnotemark[2]\\
  \and
  Cordelia Schmid\\
  Inria\footnotemark[2]\\
}

\maketitle
\ificcvfinal\thispagestyle{empty}\fi

\blfootnote{Code: \url{https://github.com/rstrudel/segmenter}}

\begin{abstract}
Image segmentation is often ambiguous at the level of individual image patches
and requires contextual information to reach label consensus.  In this paper we
introduce Segmenter,  a transformer model for semantic segmentation. In contrast
to convolution-based methods, our approach allows to model global context
already at the first layer and throughout the network. We build on the recent
Vision Transformer (ViT) and extend it to semantic
segmentation. To do so, we rely on the output embeddings corresponding to image
patches and obtain class labels from these embeddings with a point-wise linear
decoder or a mask transformer decoder. We leverage models pre-trained for image
classification and show that we can fine-tune them on moderate sized datasets available for semantic
segmentation. The linear decoder allows to obtain excellent results already, but
the performance can be further improved by a mask transformer generating class
masks.  We conduct an extensive ablation study to show the impact of the different
parameters, in particular the performance is better for large models and small
patch sizes. Segmenter attains excellent results for semantic segmentation. It
outperforms the state of the art on both ADE20K and Pascal Context
datasets and is competitive on Cityscapes.
\end{abstract}

\section{Introduction}

\begin{figure}[t]
  \centering
  \includegraphics[width=0.85\linewidth]{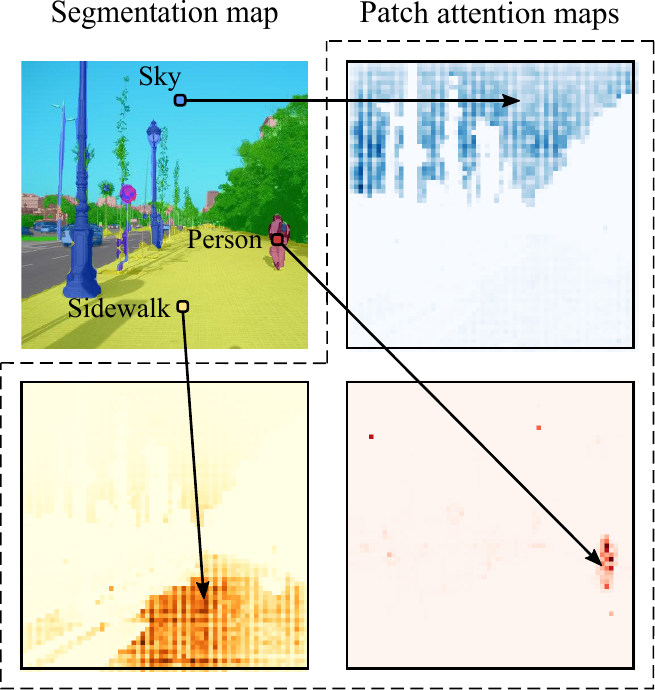}
  \caption{
    Our approach for semantic segmentation is purely transformer based. It leverages the global image context at every layer of the model.
    Attention maps from the first Segmenter layer are displayed for three
    $8 \times 8$ patches and highlight the early grouping of patches into semantically meaningful categories.
    The original image (top-left) is overlayed with segmentation masks produced by our method.
    }
  \label{fig:teaser}
  \vspace{-0.5cm}
\end{figure}

Semantic segmentation is a challenging computer vision problem with a wide range
of applications including autonomous driving, robotics, augmented reality, image
editing, medical imaging and many others~\cite{hesamian19, hong18, siam17}.  The
goal of semantic segmentation is to assign each image pixel to a category
label corresponding to the underlying object and to provide high-level image
representations for target tasks, e.g.~detecting the boundaries of people and their
clothes for virtual try-on applications~\cite{hsieh2019fashionon}.  Despite much
effort and large progress over recent years
\cite{deeplabv3plus18,fu20,huang19,minaee21,sultana20,zhao17,zhao18}, image
segmentation remains a challenging problem due to rich intra-class variation,
context variation and ambiguities originating from occlusions and low image
resolution.

Recent approaches to semantic segmentation typically rely on convolutional
encoder-decoder architectures where the encoder generates low-resolution image
features and the decoder upsamples features to segmentation maps with per-pixel
class scores.  State-of-the-art methods deploy Fully Convolutional Networks
(FCN)~\cite{long17} and achieve impressive results on challenging
segmentation benchmarks~\cite{deeplabv3plus18, fu19dual,  yin20, yu20cpn,
yuan20, zhang19, zhao18}.  These methods rely on learnable stacked convolutions
that can capture  semantically rich information and have been highly successful
in computer vision.  The local nature of convolutional filters, however, limits
the access to the global information in the image. Meanwhile, such information
is particularly important for segmentation where the labeling of local patches
often depends on the global image context.  To circumvent this issue, DeepLab
methods~\cite{deeplab18, deeplabv317,deeplabv3plus18} introduce feature
aggregation with dilated convolutions and spatial pyramid pooling.  This allows
to enlarge the receptive fields of convolutional networks and to obtain
multi-scale features.  Following recent progresses in NLP~\cite{vaswani17},
several segmentation methods explore alternative aggregation schemes based on
channel or spatial \cite{fu20, fu19dual, yuan18} attention and
point-wise~\cite{zhao18} attention to better capture contextual information.
Such methods, however, still rely on convolutional backbones and are, hence,
biased towards local interactions.  An extensive use of specialised layers to
remedy this bias~\cite{deeplab18, deeplabv3plus18, fu20, yu20cpn} suggests
limitations of convolutional architectures for segmentation.

To overcome these limitations, we formulate the problem of semantic segmentation
as a sequence-to-sequence problem and use a transformer
architecture~\cite{vaswani17} to leverage contextual information at every stage
of the model.  By design, transformers can capture global interactions between
elements of a scene and have no built-in inductive prior, see
Figure~\ref{fig:teaser}. However, the modeling of global interactions comes at a
quadratic cost which makes such methods prohibitively expensive when applied to
raw image pixels~\cite{chen20}. Following the recent work on Vision Transformers
(ViT)~\cite{vit20,deit20}, we split the image into patches and treat linear
patch embeddings as input tokens for the transformer encoder.  The
contextualized sequence of tokens produced by the encoder is then upsampled by a
transformer decoder to per-pixel class scores. For decoding, we consider either
a simple point-wise linear mapping of patch embeddings to class scores or a
transformer-based decoding scheme where learnable class embeddings are processed
jointly with patch tokens to generate class masks. We conduct an extensive study
of transformers for segmentation by ablating model regularization, model size,
input patch size and its trade-off between accuracy and performance. Our
Segmenter approach attains excellent results while remaining simple, flexible and fast.
In particular,  when using large models with small input patch size the best
model reaches a mean IoU of 53.63\% on the challenging ADE20K \cite{ade20k}
dataset, surpassing all previous state-of-the-art convolutional approaches by a
large margin of 5.3\%. Such improvement partly stems from the global context
captured by our method at every stage of the model as highlighted in Figure \ref{fig:teaser}.

In summary, our work provides the following four contributions: (i)~We propose a
novel approach to semantic segmentation based on the Vision Transformer (ViT)
\cite{vit20} that does not use convolutions, captures contextual information by
design and outperforms FCN based approaches.  (ii)~We present a family of models
with varying levels of resolution which allows to trade-off between precision
and run-time, ranging from \textit{state-of-the-art} performance to models with
fast inference and good performances.  (iii)~We propose a transformer-based
decoder generating class masks which outperforms our linear baseline and can be
extended to perform more general image segmentation tasks.   (iv)~We demonstrate
that our approach yields \textit{state-of-the-art} results on both
ADE20K \cite{ade20k} and Pascal Context \cite{pascal_context} datasets
and is competitive on Cityscapes \cite{cityscapes}.

\section{Related work}
\begin{figure*}[t]
  \centering
  \vspace{-0.2cm}
  \includegraphics[width=\textwidth]{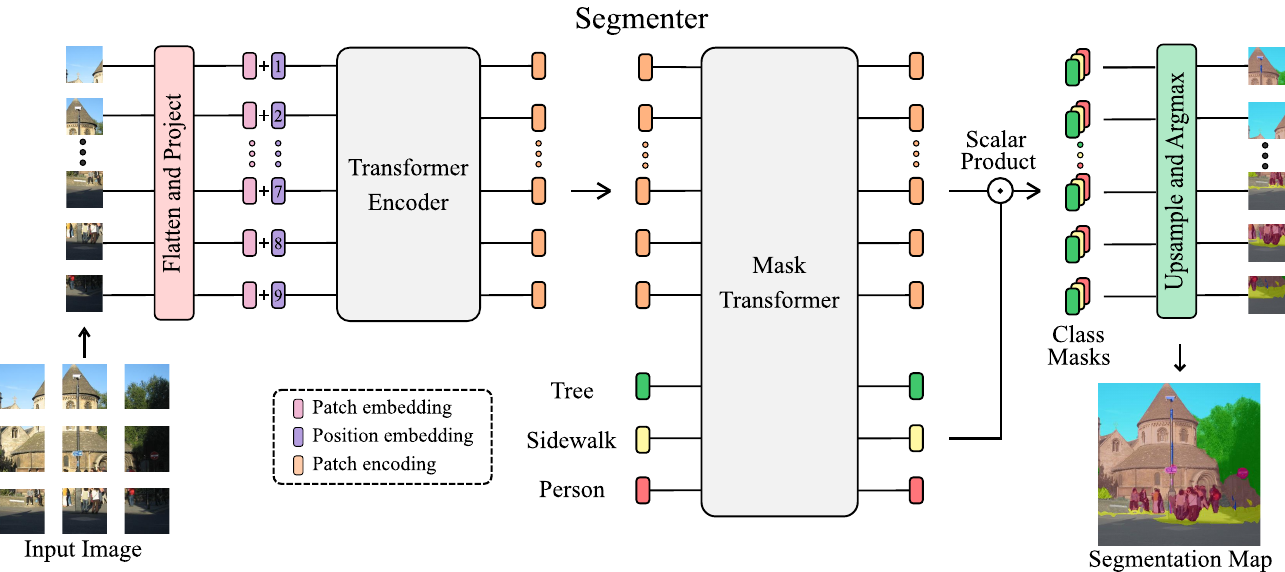}
  \caption{Overview of our approach Segmenter. (Left) Encoder: The image patches are projected to a sequence of embeddings and then encoded with a transformer. (Right)  Decoder: A mask transformer takes as input the output of the encoder and class embeddings to predict segmentation masks. See text for details.} 
  \label{fig:overview}
  \vspace{-0.5cm}
\end{figure*}

\paragraph{Semantic segmentation.}
Methods based on Fully Convolutional Networks (FCN) combined with
encoder-decoder architectures  have become the dominant approach to semantic
segmentation. Initial approaches \cite{farabet13, long15, Ning05, pinheiro14}
rely on a stack of consecutive convolutions followed by spatial pooling to
perform dense predictions. Consecutive approaches~\cite{islam17, badrinarayanan17, lin17,
pohlen17, ronnenberger15} upsample high-level feature maps and combine them with
low-level feature maps during decoding to both capture global information and
recover sharp object boundaries. To enlarge the receptive field of convolutions
in the first layers, several approaches~\cite{deeplab18, deeplab15, yu16} have proposed dilated or
atrous convolutions.  To capture global information in higher layers, recent
work~\cite{deeplabv317, deeplabv3plus18, zhao17} employs spatial pyramid pooling to
capture multi-scale contextual information. Combining these enhancements along
with atrous spatial pyramid pooling, Deeplabv3+ \cite{deeplabv3plus18} proposes
a simple and effective encoder-decoder FCN architecture. Recent work
\cite{fu20, fu19dual, yin20, yu20cpn, yuan18, zhao18} replace coarse pooling by
attention mechanisms on top of the encoder feature maps to better capture
long-range dependencies.

While recent segmentation methods are mostly focused on improving FCN, the
restriction to local operations imposed by convolutions may imply inefficient
processing of global image context and suboptimal segmentation results.  Hence, we propose a pure transformer architecture that captures global context at
every layer of the model during the encoding and decoding stages.

\vspace{-.4cm}
\paragraph{Transformers for vision.} Transformers \cite{vaswani17} are now state
of the art in many Natural Language Processing (NLP) tasks. Such  models rely on
self-attention mechanisms and capture long-range dependencies among tokens
(words) in a sentence.  In addition, transformers are well suited for
parallelization, facilitating training on large datasets.  The success of transformers
in NLP has inspired several methods in computer vision combining CNNs with forms
of self-attention to address object detection~\cite{carion20}, semantic
segmentation \cite{wang-axial20}, panoptic segmentation \cite{hwang20}, video
processing \cite{xwang18} and few-shot classification \cite{doersch20}. 

Recently, the Vision Transformer (ViT) \cite{vit20} introduced a convolution-free
transformer architecture for image classification where input images are
processed as sequences of patch tokens.  While ViT requires training on very
large datasets, DeiT \cite{deit20} proposes a token-based distillation strategy
and obtains a competitive vision transformer trained on the ImageNet-1k
\cite{deng09} dataset using a CNN as a teacher. 
Concurrent work extends this work to video classification~\cite{vivit21, bertasius21} and semantic segmentation~\cite{swin21, setr20}. In more detail, SETR \cite{setr20} uses a ViT backbone and  a standard CNN decoder. 
Swin Transformer \cite{swin21} uses a variant of ViT, composed of
local windows, shifted between layers and Upper-Net as a pyramid FCN decoder.

Here, we propose Segmenter, a transformer encoder-decoder architecture for semantic
image segmentation.  Our approach relies on a ViT backbone and introduces a mask decoder inspired by DETR \cite{carion20}.
Our architecture does not use convolutions, captures
global image context by design and results in competitive performance on
standard image segmentation benchmarks.

\section{Our approach: Segmenter}

Segmenter is based on a fully transformer-based encoder-decoder architecture
mapping a sequence of patch embeddings to pixel-level class annotations. An
overview of the model is shown in Figure \ref{fig:overview}. The sequence of
patches is encoded by a transformer encoder described in Section
\ref{sec:encoder} and decoded by either a point-wise linear mapping or a mask
transformer described in Section \ref{sec:decoder}.  Our model is trained
end-to-end with a per-pixel cross-entropy loss. At inference time, argmax is
applied after upsampling to obtain a single class per pixel.

\subsection{Encoder}
\label{sec:encoder}

An image $\mbf{x} \in \mathbb{R}^{H \times W \times C}$ is split into a sequence
of patches $\mbf{x} = [x_1, ..., x_N] \in \mathbb{R}^{N \times P^2 \times C}$
where $(P, P)$ is the patch size, $N = HW/P^2$ is the number of patches and $C$
is the number of channels.  Each patch is flattened into a 1D vector and then
linearly projected to a patch embedding to produce a sequence of patch
embeddings
$\mbf{x_0} = [\mbf{E}x_1, ..., \mbf{E}x_N] \in \mathbb{R}^{N \times D}$ where
$\mbf{E} \in \mathbb{R}^{D \times (P^2C)}$. To capture positional information,
learnable position embeddings
$\mbf{pos} = [\text{pos}_1, ..., \text{pos}_N] \in \mathbb{R}^{N \times D}$ are
added to the sequence of patches to get the resulting input sequence of tokens
$\mbf{z_0} = \mbf{x_0}+\mbf{pos}$.

A transformer \cite{vaswani17} encoder composed of $L$ layers is applied to the
sequence of tokens $\mbf{z}_0$ to generate a sequence of contextualized
encodings $\mbf{z}_L \in \mathbb{R}^{N \times D}$. A transformer layer consists
of a multi-headed self-attention (MSA) block followed by a point-wise MLP block
of two layers with layer norm (LN) applied before every block and residual
connections added after every block:
\begin{eqnarray} \mbf{a_{i-1}} &=& \text{MSA}(\text{LN}(\mbf{z_{i-1}})) + \mbf{z_{i-1}},\\ \mbf{z_{i}} &=& \text{MLP}(\text{LN}(\mbf{a_{i-1}})) + \mbf{a_{i-1}},
\end{eqnarray} where $i \in \{1, ..., L\}$. The self-attention mechanism is
composed of three point-wise linear layers mapping tokens to intermediate
representations, queries $\mbf{Q} \in \mathbb{R}^{N \times d}$, keys
$\mbf{K} \in \mathbb{R}^{N \times d}$ and values
$\mbf{V} \in \mathbb{R}^{N \times d}$.  Self-attention is then computed as
follows
\begin{equation}
  \text{MSA}(\mbf{Q}, \mbf{K}, \mbf{V}) = \text{softmax}\left(\frac{\mbf{QK}^T}{\sqrt{d}}\right)\mbf{V}.
\end{equation}

The transformer encoder maps the input sequence
$\mbf{z_{0}}=[z_{0,1}, ..., z_{0,N}]$ of embedded patches with position encoding
to  $\mbf{z_{L}}=[z_{L,1}, ..., z_{L,N}]$, a contextualized encoding sequence
containing rich semantic information used by the decoder. In the following
section we introduce the decoder.

\subsection{Decoder}
\label{sec:decoder}

The sequence of patch encodings $\mbf{z_L} \in \mathbb{R}^{N \times D}$ is
decoded to a segmentation map $\mbf{s} \in \mathbb{R}^{H \times W \times K}$ where $K$
is the number of classes. The decoder learns to map patch-level encodings coming
from the encoder to patch-level class scores. Next these patch-level class
scores are upsampled by bilinear interpolation to pixel-level scores. We describe in the following a
linear decoder, which serves as a baseline, and our approach, a mask transformer, see Figure~\ref{fig:overview}.

\noindent \textbf{Linear.}  A point-wise linear layer is applied to the patch
encodings $\mbf{z_L} \in \mathbb{R}^{N \times D}$ to produce patch-level class
logits $\mbf{z_{lin}} \in \mathbb{R}^{N \times K}$. The sequence is then
reshaped into a 2D feature map $\mbf{s_{lin}} \in \mathbb{R}^{H/P \times W/P \times K}$
and bilinearly upsampled to the original image size
$\mbf{s} \in \mathbb{R}^{H \times W \times K}$. A softmax is then applied on the
class dimension to obtain the final segmentation map.

\noindent \textbf{Mask Transformer.} For the transformer-based decoder, we
introduce a set of $K$ learnable class embeddings
$\mbf{cls} = [\text{cls}_1, ..., \text{cls}_K] \in \mathbb{R}^{K \times D}$
where $K$ is the number of classes.  Each class embedding is initialized
randomly and assigned to a single semantic class. It will be used to generate
the class mask. The class embeddings $\mbf{cls}$ are processed jointly with
patch encodings $\mbf{z_L}$ by the decoder as depicted in Figure
\ref{fig:overview}. The decoder is a transformer encoder composed of $M$ layers.
Our mask transformer generates $K$ masks by computing the scalar product between
L2-normalized patch embeddings $\mbf{z_{M}'} \in \mathbb{R}^{N\times D}$ and class
embeddings $\mbf{c} \in \mathbb{R}^{K\times D}$ output by the decoder.  The set
of class masks is computed as follows
\begin{equation}
  \text{Masks}(\mbf{z_{M}'}, \mbf{c}) = \mbf{z_{M}'c}^T
\end{equation}
where $\text{Masks}(\mbf{z_{M}'}, \mbf{c}) \in \mathbb{R}^{N \times K}$ is a
set of patch sequence. Each mask sequence is then reshaped into a 2D mask to form
$\mbf{s_{mask}} \in \mathbb{R}^{H/P \times W/P \times K}$ and bilinearly
upsampled to the original image size to obtain a feature map
$\mbf{s} \in \mathbb{R}^{H \times W \times K}$. A softmax is then applied on the
class dimension followed by a layer norm to obtain pixel-wise class score
forming the final segmentation map. The masks sequences are softly exclusive to
each other i.e. $\sum_{k=1}^{K} s_{i, j, k} = 1$ for all $(i, j) \in H\times W$.

Our mask transformer is inspired by DETR \cite{carion20}, Max-DeepLab \cite{wang-max20} and SOLO-v2 \cite{xwang20} which introduce object embeddings \cite{carion20} to produce instance masks \cite{wang-max20, xwang20}.  However, unlike our method, MaxDeep-Lab uses an hybrid approach
based on CNNs and transformers and splits the pixel and class embeddings into
two streams because of computational constraints. Using a pure transformer
architecture and leveraging  patch level encodings, we propose a simple approach
that processes the patch and class embeddings jointly during the decoding phase.
Such approach allows to produce dynamical filters, changing with the input.
While we address semantic segmentation in this work, our mask transformer can also be
directly adapted to perform panoptic segmentation by replacing the class
embeddings by object embeddings.

\section{Experimental results}

\subsection{Datasets and metrics}

\noindent \textbf{ADE20K}~\cite{ade20k}. This dataset contains challenging scenes
with fine-grained labels and is one of the most challenging semantic
segmentation datasets.
The training set contains 20,210 images with 150 semantic classes. The validation and test set contain
2,000 and 3,352 images respectively. \\
\textbf{Pascal Context} \cite{pascal_context}. The
training set  contains 4,996 images  with 59 semantic classes plus a
background class. The validation set contains 5,104 images.\\
\textbf{Cityscapes}~\cite{cityscapes}. The dataset contains 5,000 images from 50
different cities with 19 semantic classes. There are 2,975 images in the training
set, 500 images in the validation set and 1,525 images in the test set.

\noindent \textbf{Metrics.} We report Intersection over Union (mIoU) averaged
over all classes.

\begin{figure*}[t]

\centering
\centering
  \vspace{-0.4cm}
\begin{subfigure}[b]{0.22\textwidth}
    \includegraphics[width=\textwidth]{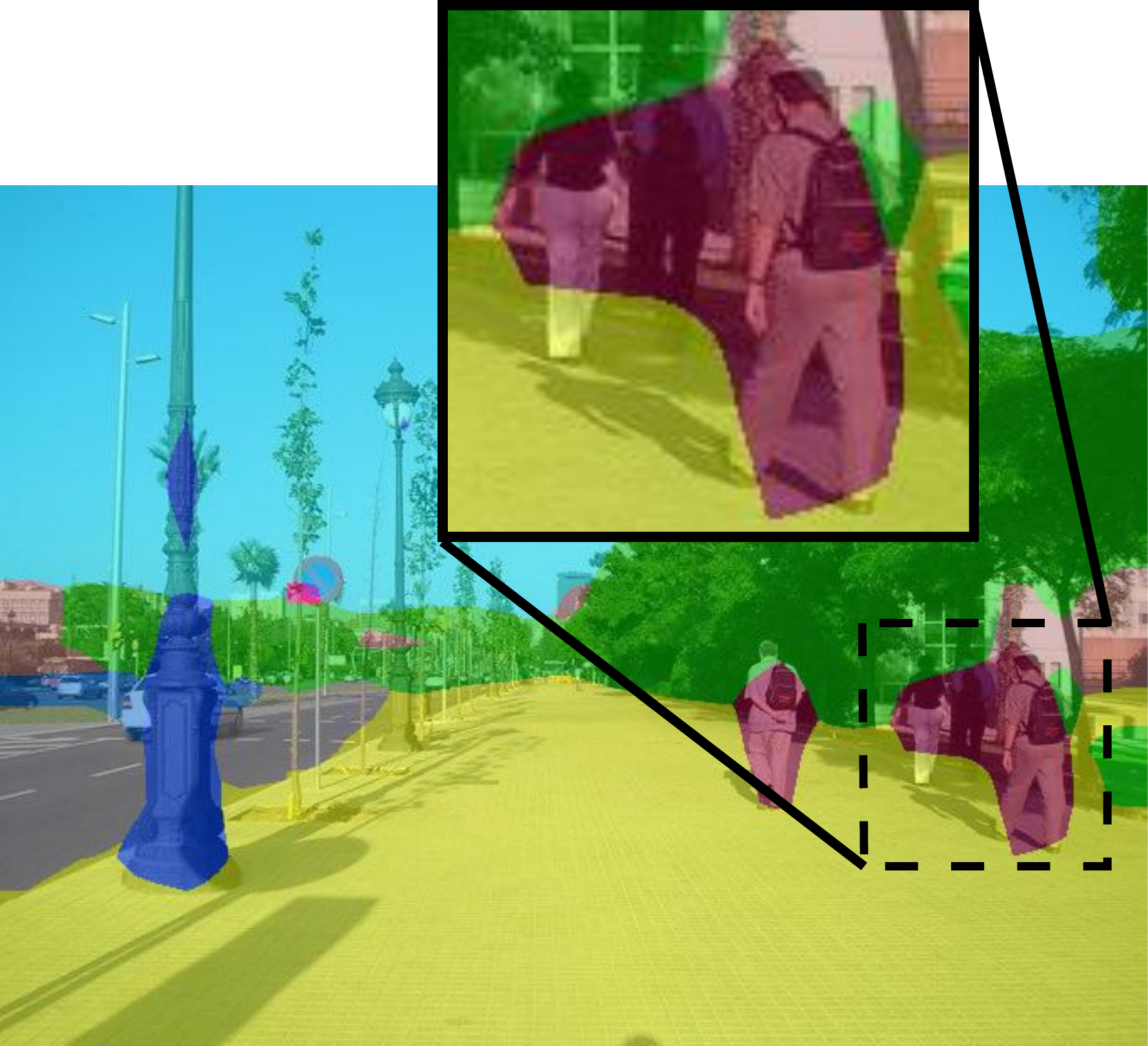}
    \caption{Patch size $32 \times 32$}
\end{subfigure}
\begin{subfigure}[b]{0.22\textwidth}
    \includegraphics[width=\textwidth]{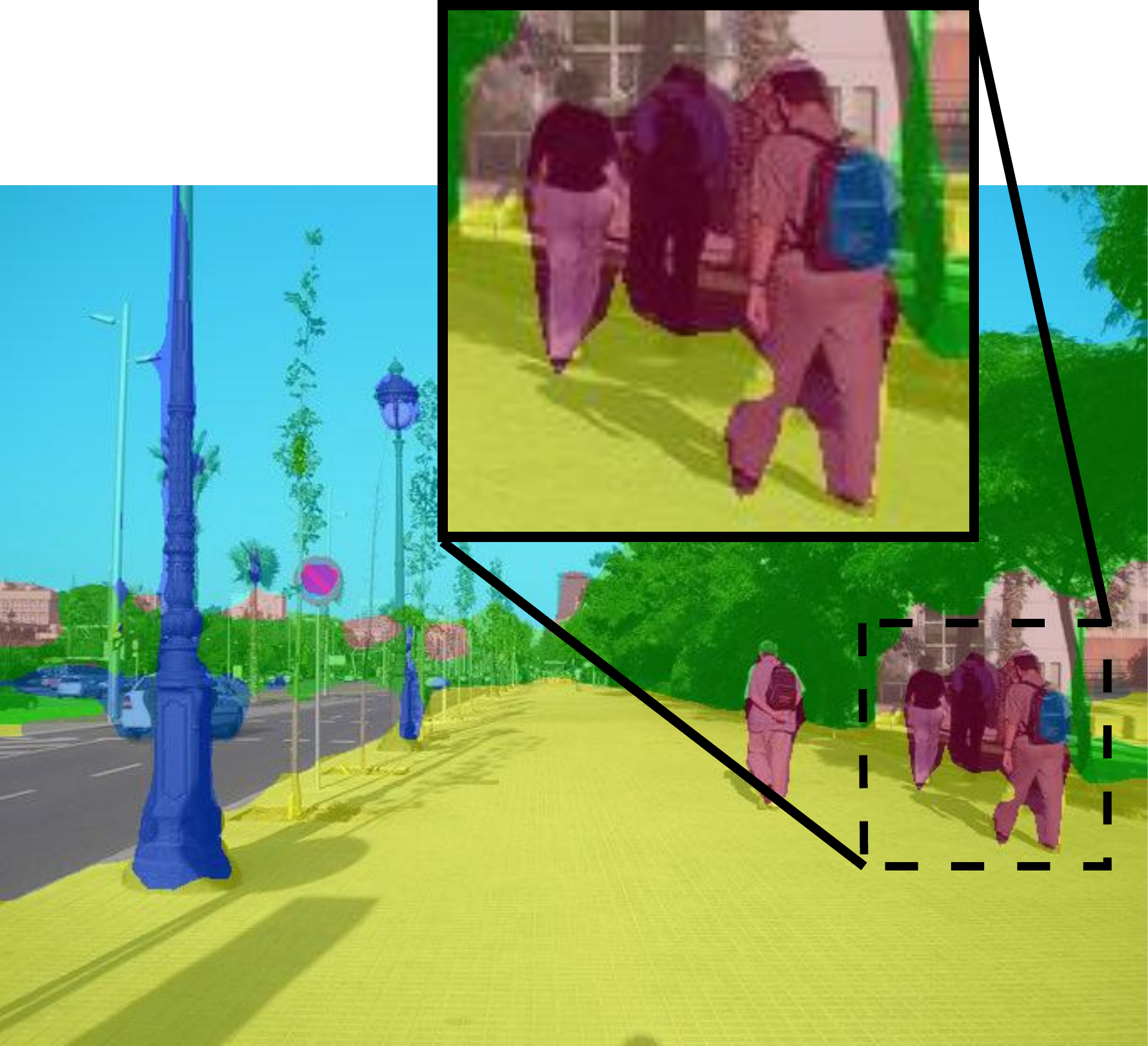}
    \caption{Patch size $16 \times 16$}
\end{subfigure}
\begin{subfigure}[b]{0.22\textwidth}
    \includegraphics[width=\textwidth]{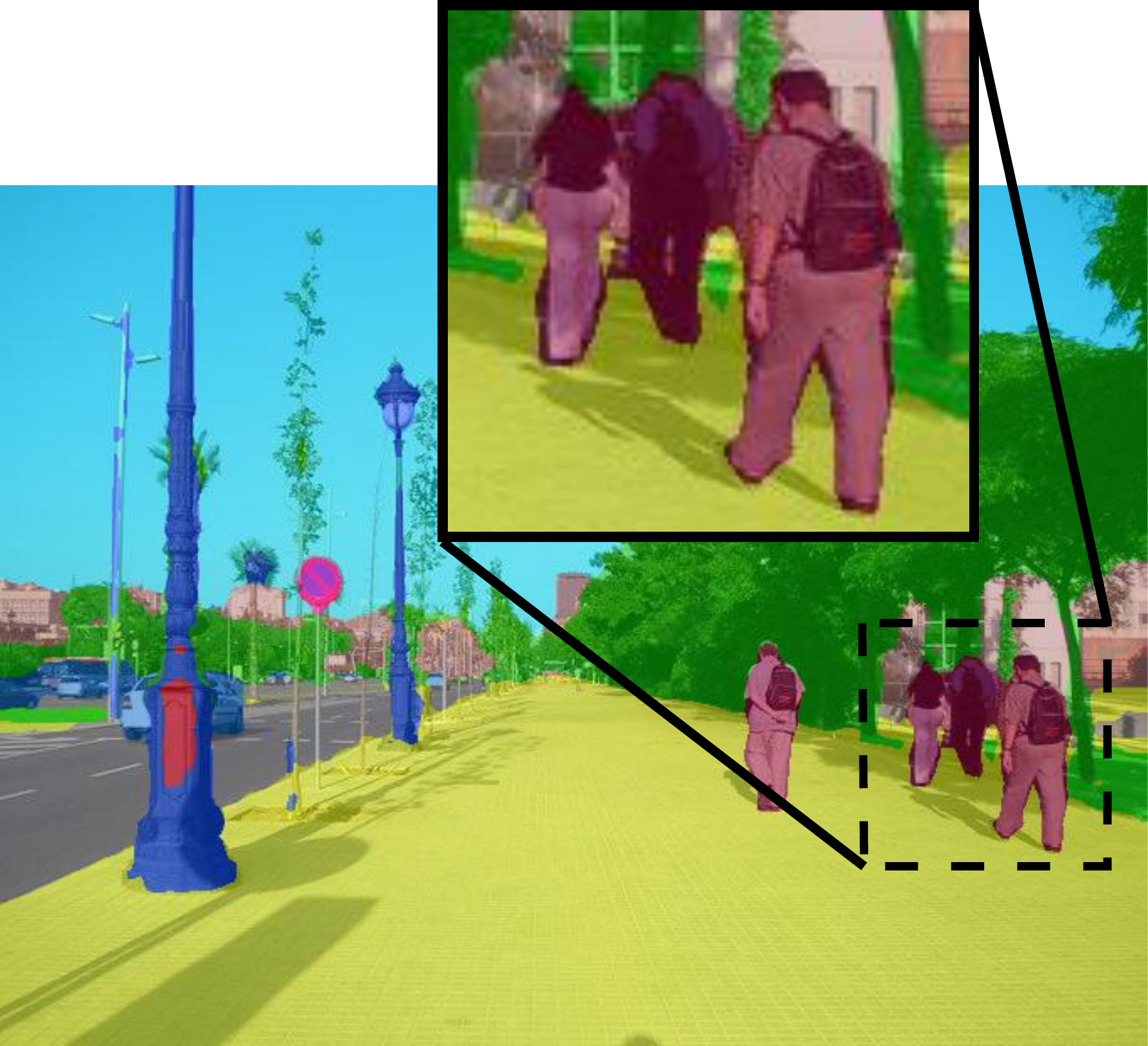}
    \caption{Patch size $8 \times 8$}
\end{subfigure}
\begin{subfigure}[b]{0.22\textwidth}
    \includegraphics[width=\textwidth]{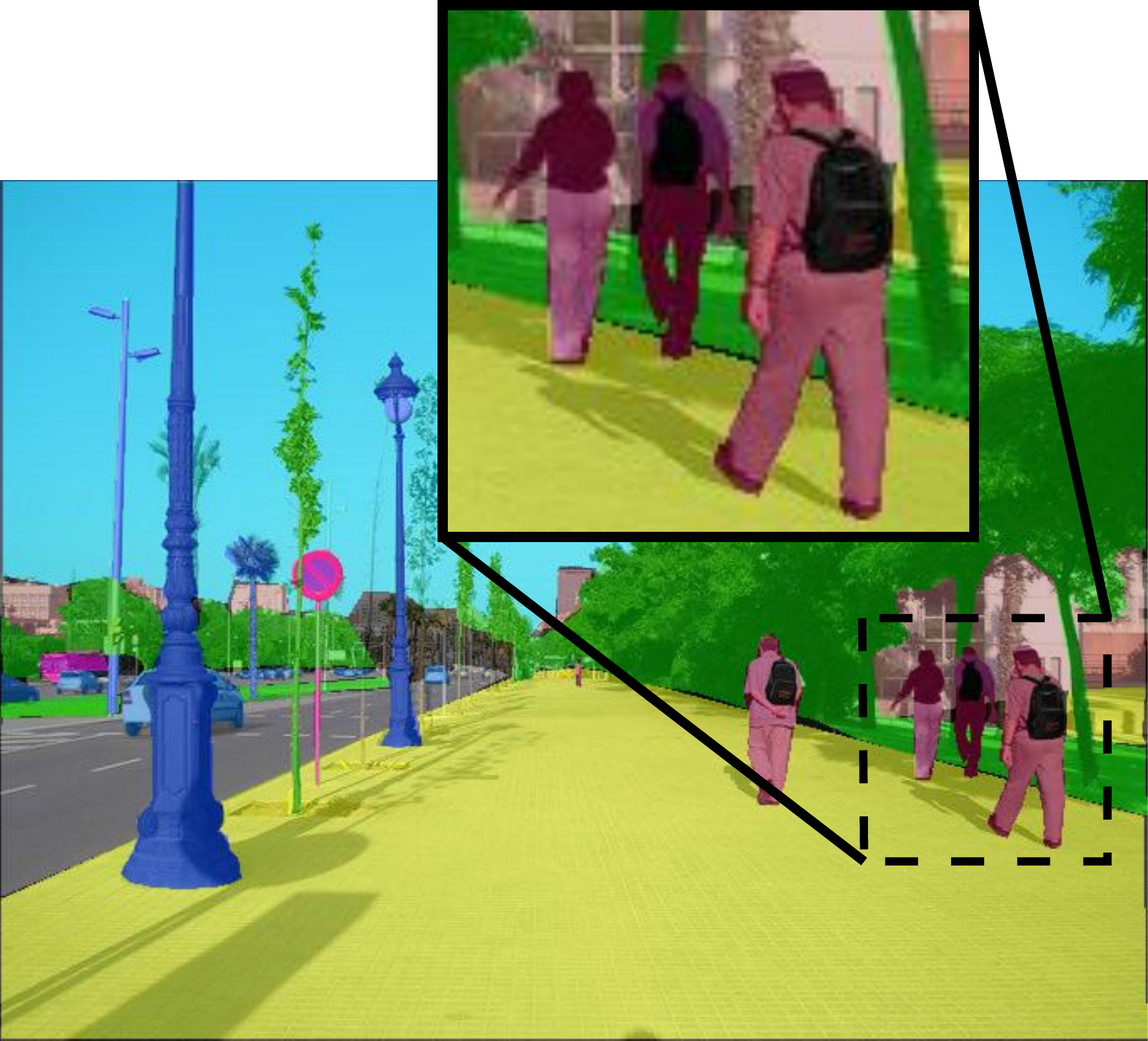}
    \caption{Ground Truth}
\end{subfigure}
\vspace{-0.2cm}
\caption{Impact of the model patch size on the segmentation maps.}
\label{fig:patch_size}
\vspace{-0.4cm}

\end{figure*}

\begin{table}
\small
\setlength\tabcolsep{2.7pt}
\centering
\begin{tabular}{llccccc}
\toprule
Model & Backbone & Layers & Token size & Heads & Params \\
\midrule
Seg-Ti & ViT-Ti & 12 & 192 & 3 & 6M \\
Seg-S & ViT-S & 12 & 384 & 6 & 22M \\
Seg-B & ViT-B & 12 & 768 & 12 & 86M \\
Seg-B\tsdagger & DeiT-B & 12 & 768 & 12 & 86M \\
Seg-L & ViT-L & 24 & 1024 & 16 & 307M \\
\bottomrule
\end{tabular}
\vspace{-0.2cm}
\caption{Details of Transformer variants.}
\label{table:desc}
\vspace{-0.6cm}
\end{table}

\subsection{Implementation details}

\noindent \textbf{Transformer models.} For the encoder, we build upon the vision
transformer ViT \cite{vit20}
and consider ''Tiny'', ''Small'', ''Base'' and ''Large''
models described in Table \ref{table:desc}. 
The parameters varying in the
transformer encoder are the number of layers and the token size. The head size
of a multi-headed self-attention (MSA) block is fixed to 64, the number of heads
is the token size divided by the head size and the hidden size of the MLP
following MSA is four times the token size.
We also use DeiT~\cite{deit20}, a variant of the 
vision transformer. We consider models representing the image at different resolutions and use input patch sizes 
$8\!\times\!8$, $16\!\times\!16$ and $32\!\times\!32$.
In the following, we use an abbreviation to describe the model variant and patch size, 
for instance Seg-B/16 denotes the ''Base'' variant with $16\!\times\!16$ input patch size. Models based on
DeiT are denoted with a \tsdagger, for instance Seg-B\tsdagger/16.

\noindent \textbf{ImageNet pre-training.} Our Segmenter models are
pre-trained on ImageNet, ViT is pre-trained on ImageNet-21k with strong data
augmentation and regularization \cite{steiner21} and its variant DeiT is
pre-trained on ImageNet-1k. The original ViT models \cite{vit20} have been
trained with random cropping only, whereas the training procedure proposed by
\cite{steiner21} uses a combination of dropout \cite{srivastava14} and stochastic depth \cite{huang16} as regularization and Mixup \cite{mixup18} and RandAugment \cite{randaugment20} as data augmentations. This significantly improves the ImageNet top-1 accuracy, i.e., it obtains a gain of +2\% on ViT-B/16. We fine-tuned ViT-B/16 on ADE20K with models from~\cite{vit20} and~\cite{steiner21} and observe a significant difference, namely a mIoU of 45.69\% and 48.06\% respectively. In the following, all the Segmenter models will be
initialized with the improved ViT models from \cite{steiner21}.  We use
publicly available models provided by the image classification library timm
\cite{timm20} and Google research~\cite{vit_models}.  Both models are
pre-trained at an image resolution of 224 and fine-tuned on ImageNet-1k at a
resolution of 384, except for ViT-B/8 which has been fine-tuned at a
resolution of 224. We keep the patch size fixed and fine-tune the models for
the semantic segmentation task at higher resolution depending on the dataset. As
the patch size is fixed, increasing resolution results in longer token
sequences.  Following~\cite{vit20}, we bilinearly interpolate the pre-trained
position embeddings according to their original position in the image to match
the fine-tuning sequence length.  The decoders, described in Section
\ref{sec:decoder} are initialized with random weights from a truncated normal
distribution \cite{hanin18}.

\noindent \textbf{Data augmentation.} During training, we follow the standard
pipeline from the semantic segmentation library
MMSegmentation \cite{mmseg20}, which does mean substraction,
random resizing of the image to a ratio between 0.5 and 2.0 and random
left-right flipping. We randomly crop large images and pad small images
to a fixed size of $512\!\times\!512$ for ADE20K, $480\!\times\!480$ for
Pascal-Context and $768\!\times\!768$ for Cityscapes. On ADE20K, we train our largest model Seg-L-Mask/16 with a resolution of $640\!\times\!640$, matching the resolution used by the Swin Transformer \cite{swin21}.

\noindent \textbf{Optimization.} To fine-tune the pre-trained models for the
semantic segmentation task, we use the standard pixel-wise cross-entropy loss
without weight rebalancing. We use stochastic gradient descent (SGD)
\cite{robbins1951} as the optimizer with a base learning rate $\gamma_{0}$ and
set weight decay to 0. Following the seminal work of DeepLab \cite{deeplab15} we
adopt the ''poly'' learning rate decay
$\gamma = \gamma_0\,(1-\frac{N_{iter}}{N_{total}})^{0.9}$ where $N_{iter}$ and
$N_{total}$ represent the current iteration number and the total iteration
number. For ADE20K, we set the base learning rate $\gamma_0$ to $10^{-3}$ and
train for 160K iterations with a batch size of 8. For Pascal Context, we set
$\gamma_{0}$ to $10^{-3}$ and train for 80K iterations with a batch size of 16.
For Cityscapes, we set $\gamma_0$ to $10^{-2}$ and train for 80K iterations with
a batch size of 8. The schedule is similar to DeepLabv3+ \cite{deeplabv3plus18}
with learning rates divided by a factor 10 except for Cityscapes where we use a
factor of 1.

\noindent \textbf{Inference.} To handle varying image sizes during inference,  we use a
sliding-window with a resolution matching the training size. For multi-scale
inference, following standard practice \cite{deeplabv3plus18} we use rescaled
versions of the image with scaling factors of (0.5, 0.75, 1.0, 1.25, 1.5, 1.75)
and left-right flipping and average the results.

\subsection{Ablation study} 

In this section, we ablate different variants of our approach on the ADE20K
validation set. We investigate model regularization, model size, patch size, model
performance, training dataset size, compare Segmenter to convolutional
approaches and evaluate different decoders. Unless stated otherwise, we use the
baseline linear decoder and report results using single-scale inference.

\begin{table}
\small
\setlength\tabcolsep{2.7pt} 
\centering
\begin{tabular}{ccccc}
\toprule
  & & \multicolumn{3}{c}{Stochastic Depth}\\
  & & 0.0 & 0.1 & 0.2 \\
  \midrule
  \parbox[t]{3mm}{\multirow{3}{*}{\rotatebox[origin=c]{90}{Dropout}}}
  & 0.0 \vrule & 45.01 & \textbf{45.37} & 45.10 \\
  & 0.1 \vrule & 42.02 & 42.30 & 41.14 \\
  & 0.2 \vrule & 36.49 & 36.63 & 35.67 \\
\bottomrule
\end{tabular}
\vspace{-0.2cm}
\caption{Mean IoU comparison of different regularization schemes using Seg-S/16 on ADE20K validation set.}
\label{table:regularization}
\vspace{-0.1cm}
\end{table}


\begin{table}
\small
\setlength\tabcolsep{2.7pt}
\centering
\begin{tabular}{llcccc}
  \toprule
  Method & Backbone & Patch size & Im/sec & ImNet acc. & mIoU (SS) \\
  \midrule
  Seg-Ti/16 & ViT-Ti & 16 & 396 & 78.6 & 39.03 \\
  \midrule
  Seg-S/32 & ViT-S & 32 & 1032 & 80.5 & 40.64 \\
  Seg-S/16 & ViT-S & 16 & 196 & 83.7 & 45.37 \\
  \midrule
  Seg-B\tsdagger/16 & DeiT-B & 16 & 92 & 85.2 & 47.08 \\
  \midrule
  Seg-B/32 & ViT-B & 32 & 516 & 83.3 & 43.07 \\
  Seg-B/16 & ViT-B & 16 & 92 & 86.0 & 48.06 \\
  Seg-B/8 & ViT-B & 8 & 7 & 85.7 & 49.54 \\
  \midrule
  Seg-L/16 & ViT-L & 16 & 33 & 87.1 & 50.71 \\
\bottomrule
\end{tabular}
\vspace{-0.2cm}
\caption{Performance comparison of different Segmenter models with varying backbones and input
  patch sizes on ADE20K validation set.}
\label{table:models}
\vspace{-0.5cm}
\end{table}


\noindent
\textbf{Regularization.} We first compare two forms of regularization, dropout
\cite{srivastava14} and stochastic depth \cite{huang16}, and show that
stochastic depth consistently improves transformer training for segmentation.
CNN models rely on batch normalization~\cite{ioffe15} which also acts as a
regularizer. In contrast, transformers are usually composed of layer
normalization~\cite{ba16} combined with dropout as a regularizer during training~\cite{devlin19, vit20}. Dropout randomly ignores tokens given as input of a
block and stochastic depth randomly skips a learnable block of the model during
the forward pass. We compare regularizations on Seg-S/16 based
on ViT-S/16 backbone. Table \ref{table:regularization}
shows that stochastic depth set to 0.1, dropping 10\% of the layers randomly,
consistently improves the performance, with 0.36\% when the dropout is set to 0
compared to the baseline without regularization.  Dropout consistently hurts
performances, either alone or when combined with stochastic depth. This is
consistent with \cite{deit20} which observed the negative impact of dropout for
image classification. From now on, all the models will be trained with
stochastic depth set to 0.1 and without dropout.

\noindent
\textbf{Transformer size.} We now study the impact of transformers size on
performance by varying the number of layers and the tokens size for a fixed
patch size of 16. Table \ref{table:models} shows that performance scales
nicely with the backbone capacity. When doubling the token dimension,
from Seg-S/16 to Seg-B/16, we get a 2.69\% improvement. When doubling the number
of layers, from Seg-B/16 to Seg-L/16, we get an improvement of 2.65\%.
Finally, our largest Segmenter model, Seg-L/16, achieves a strong mIoU of 50.71\% with a simple decoding scheme on the ADE20K validation dataset with single scale
inference. The absence of tasks-specific layers vastly used in FCN models
suggests that transformer based methods provide more expressive models, well
suited for semantic segmentation.

\begin{figure}[t]
  \centering
  \includegraphics[width=0.95\linewidth]{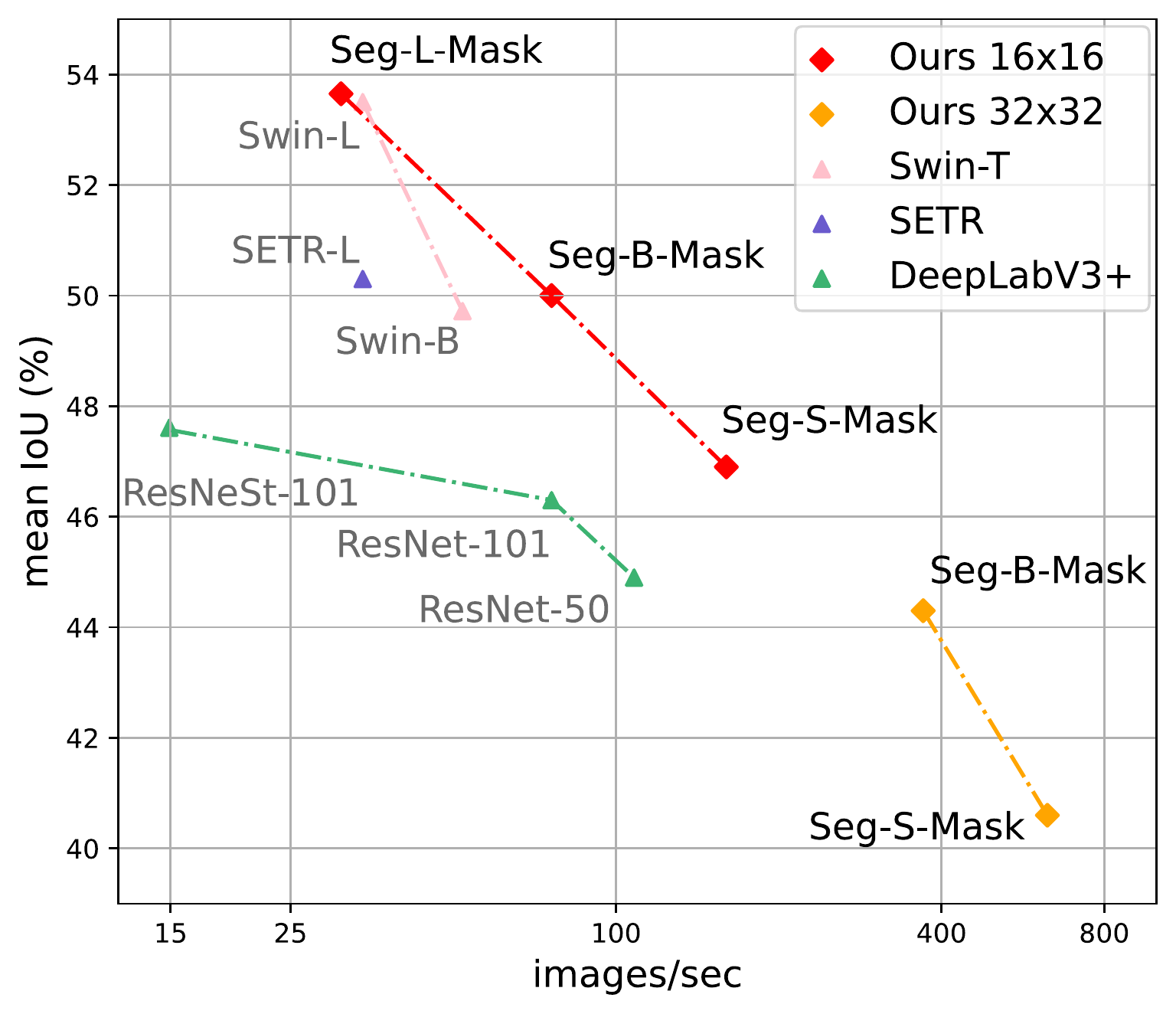}
  \vspace{-0.4cm}
  \caption{Images per second and mean IoU for our approach
    compared to other methods on ADE20K validation set. Segmenter models offer a competitive trade-off in
    terms of performance and precision.}
    \vspace{-0.5cm}
  \label{fig:performance}
\end{figure}

\begin{table}
\small
\setlength\tabcolsep{2.7pt}
\centering
\begin{tabular}{llcccc}
  \toprule
Method & Decoder & Small & Medium & Large & mIoU (SS) \\
\midrule
DeepLab RNeSt-101 & UNet & 37.85 & 50.89 & 50.67 & 46.47 \\
\midrule
Seg-B/32 & Linear & 31.95 & 47.82 & 49.44 & 43.07 \\
\rowcolor{LightCyan}Seg-B-Mask/32 & Mask & 32.29 & 49.44 & 50.82 & 44.19 \\
\midrule
Seg-B\tsdagger/16 & Linear & 38.31 & 50.91 & 52.08 & 47.10 \\
\rowcolor{LightCyan}Seg-B\tsdagger-Mask/16 & Mask & 40.49 & 51.37 & 54.24 & 48.70\\
\midrule
Seg-B/16 & Linear & 39.57 & 51.32 & 53.28 & 48.06 \\
\rowcolor{LightCyan}Seg-B-Mask/16 & Mask & 40.16 & 52.61 & 52.66 & 48.48 \\
\midrule
Seg-B/8 & Linear & 41.43 & 54.35 & 52.85 & 49.54 \\
\midrule
Seg-L/16 & Linear & 42.08 & 54.67 & 55.39 & 50.71 \\
\rowcolor{LightCyan}Seg-L-Mask/16 & Mask & 42.02 & 54.83 & 57.06 & 51.30 \\
\bottomrule
\end{tabular}
\vspace{-0.2cm}
\caption{
  Evaluation with respect to the object size on ADE20k validation set
  (mean IoU). Comparison of DeepLabv3+ ResNeSt-101 to Segmenter models with a linear or a mask
  transformer decoder.
  }
\label{table:obj_size}
\vspace{-0.1cm}
\end{table}

\begin{table}
\small
\setlength\tabcolsep{2.7pt}
\centering
\begin{tabular}{lccccc}
  \toprule
Dataset Size & 4k & 8k & 12k & 16k & 20k\\
\midrule
mIoU (SS) & 38.31 & 41.87 & 43.42 & 44.61 & 45.37\\
\bottomrule
\end{tabular}
\vspace{-0.2cm}
\caption{Performance comparison of Seg-S/16 models trained with
increasing dataset size and evaluated on ADE20K validation set.}
\label{table:dataset_size}
\vspace{-0.6cm}
\end{table}

\noindent
\textbf{Patch size.} Representing an image with a patch sequence provides a
simple way to trade-off between speed and accuracy by varying the patch size.
While increasing the patch size results in a coarser representation of the
image, it results in a smaller sequence that is faster to process. The third and
fourth parts of Table \ref{table:models} report the performance for ViT
backbones and varying patch sizes. We observe that the patch size is a key
factor for semantic segmentation performance. It is similarly important to
the model size.  Indeed, going from a patch size 32 to 16 we observe an
improvement of 5\% for Seg-B. For Seg-B, we also report
results for a patch size of 8 and report an mIoU of 49.54\%, reducing the gap
from ViT-B/8 to ViT-L/16 to 1.17\% while requiring substantially fewer parameters. This
trend shows that reducing the patch size is a robust source of improvement which
does not introduce any parameters but requires to compute attention over longer
sequences, increasing the compute time and memory footprint. If it was
computationally feasible, ViT-L/8 would probably be the best performing model.
Going towards more computation and memory efficient transformers
handling larger sequence of smaller patches is a promising direction.

To further study the impact of patch size, we show segmentation maps generated
by Segmenter models with decreasing patch size in Figure \ref{fig:patch_size}.
We observe that for a patch size of 32, the model learns a globally meaningful
segmentation but produces poor boundaries, for example the two persons on the
left are predicted by a single blob. Reducing the patch size leads to
considerably sharper boundaries as can be observed when looking at the contours
of persons. Hard to segment instances as the thin streetlight pole in the
background are only captured at a resolution of 8. In Table
\ref{table:obj_size}, we report mean IoU with respect to the object size and compare
Segmenter to DeepLabv3+ with ResNeSt backbone. To reproduce DeepLabv3+
results, we used models from the MMSegmentation library \cite{mmseg20}. We observe how Seg-B/8 improvement over Seg-B/16 comes mostly from small and medium instances with a gain of 1.27\% and 1.74\% respectively. Also, we observe that overall the biggest improvement of Segmenter over DeepLab comes from large instances where Seg-L-Mask/16 shows an improvement of 6.39\%.

\noindent\textbf{Decoder variants.}  In this section, we compare different
decoder variants. We evaluate the mask transformer introduced in Section
\ref{sec:decoder} and compare it to the linear baseline.  The mask transformer
has 2 layers with the same token and hidden size as the encoder.
Table~\ref{table:obj_size} reports the mean IoU performance. The mask transformer
provides consistent improvements over the linear baseline. The most significant
gain of 1.6\% is obtained for Seg-B\tsdagger/16, for Seg-B-Mask/32 we obtain a
1.1\% improvement and for Seg-L/16 a gain of 0.6\%. In Table
\ref{table:obj_size} we also examine the gain of different models with respect to
the object size. We observe gains both on small and large objects, showing the
benefit of using dynamical filters. In most cases the gain is more significant
for large objects, i.e., 1.4\% for Seg-B/32, 2.1\% for
Seg-B\tsdagger/16 and and 1.7\% for Seg-L/16. The class embeddings learned by the
mask transformer are semantically meaningful, i.e., similar classes are nearby,
see Figure \ref{fig:svd} for more details.

\noindent
\textbf{Transformer versus FCN.} Table \ref{table:obj_size} and Table \ref{table:ade20k}
compare our approach to FCN models and DeepLabv3+ \cite{deeplabv3plus18} with ResNeSt
backbone~\cite{resnest}, one of the best fully-convolutional approaches.  Our
transformer approach provides a significant improvement over this
state-of-the-art convolutional approach, highlighting the ability of
transformers to capture global scene understanding. Segmenter
consistently outperforms DeepLab on large instances with an improvement of more than 4\% for Seg-L/16 and 6\% for Seg-L-Mask/16. However, DeepLab performs similarly to Seg-B/16 on small and medium instances while having a similar number of parameters. Seg-B/8 and Seg-L/16 perform best on small and medium instances though at higher computational cost.

\noindent
\textbf{Performance.} In Figure \ref{fig:performance}, we compare our models to
several \textit{state-of-the-art} methods in terms of images per seconds and
mIoU and show a clear advantage of Segmenter over FCN based models (green
curve).  We also show that our approach compares favorably to recent transformer
based approach, our largest model Seg-L-Mask/16 is on-par with Swin-L and
outperforms SETR-MLA.  We observe that Seg/16
models perform best in terms of accuracy versus compute time with Seg-B-Mask/16
offering a good trade-off. Seg-B-Mask/16 outperforms FCN based approaches with
similar inference speed, matches SETR-MLA while being twice faster and requiring
less parameters and outperforms Swin-B both in terms of inference speed and
performance.  Seg/32 models learn coarser segmentation maps as discussed in the
previous section and enable fast inference with 400 images per second for
Seg-B-Mask/32, four times faster than ResNet-50 while providing similar
performances. To compute the images per second, we use a V100 GPU, fix the image
resolution to 512 and for each model we maximize the batch size allowed by
memory for a fair comparison.

\noindent
\textbf{Dataset size.} Vision Transformers highlighted the importance of large
datasets to attain good performance for the task of image classification. At the
scale of a semantic segmentation dataset, we analyze Seg-S/16 performance
on ADE20k dataset in Table \ref{table:dataset_size} when trained with a dataset
of increasing size.  We observe an important drop in performance when the
training set size is below 8k images. This shows that even during fine-tuning
transformers performs best with a sufficient amount of data.

\begin{figure*}[t]
  \centering
  \vspace{-0.6cm}
  \includegraphics[width=\linewidth]{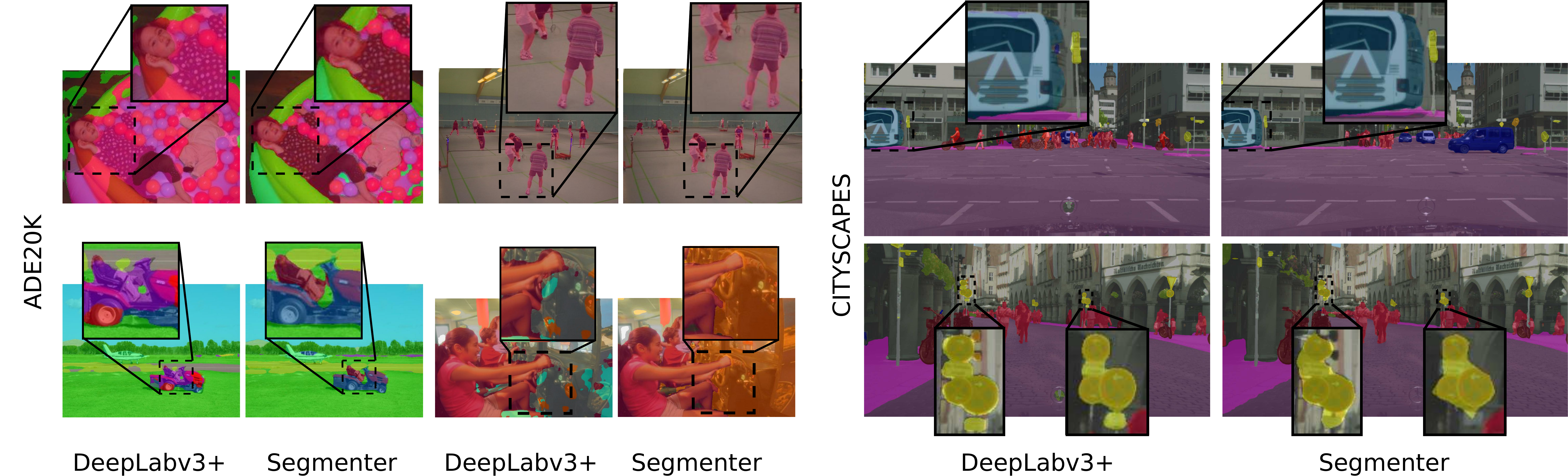}
  \vspace{-0.6cm}
  \caption{Qualitative comparison of Seg-L-Mask/16 performance with DeepLabV3+ ResNeSt-101. See Section~\ref{sec:qual} for additional qualitative results.}
  \label{fig:qualitative}
  \vspace{-0.2cm}
\end{figure*}

\begin{table}
\small
\setlength\tabcolsep{2.7pt} 
\centering
\begin{tabular}{llccc}
\toprule
Method & Backbone & Im/sec & mIoU & +MS \\
\midrule 
OCR \cite{yuan20} & HRNetV2-W48 & 83 & - & 45.66 \\
ACNet \cite{fu19adaptive} & ResNet-101 & - & - & 45.90 \\
DNL \cite{yin20} & ResNet-101 & - & - & 45.97 \\
DRANet \cite{fu20} & ResNet-101 & - & - & 46.18 \\
CPNet \cite{yu20cpn} & ResNet-101 & - & - & 46.27 \\
DeepLabv3+ \cite{deeplabv3plus18} & ResNet-101 & 76 & 45.47 & 46.35 \\
DeepLabv3+ \cite{deeplabv3plus18} & ResNeSt-101 & 15 & 46.47 & 47.27 \\
DeepLabv3+ \cite{deeplabv3plus18} & ResNeSt-200 & - & - & 48.36 \\
\midrule
SETR-L MLA \cite{setr20} & ViT-L/16 & 34 & 48.64 & 50.28 \\
Swin-L UperNet \cite{swin21} & Swin-L/16 & 34 & 52.10 & \textbf{53.50} \\
\midrule
\rowcolor{LightCyan}Seg-B\tsdagger/16 & DeiT-B/16 & 77 & 47.08 & 48.05 \\
\rowcolor{LightCyan}Seg-B\tsdagger-Mask/16 & DeiT-B/16 & 76 & 48.70 & 50.08 \\
\rowcolor{LightCyan}Seg-L/16 & ViT-L/16 & 33 & 50.71 & 52.25 \\
\rowcolor{LightCyan}Seg-L-Mask/16  & ViT-L/16 & 31 & 51.82 & \textbf{53.63} \\
\bottomrule
\end{tabular}
\vspace{-0.2cm}
\caption{State-of-the-art comparison on ADE20K validation set.}
\label{table:ade20k}
\vspace{-0.5cm}
\end{table}

\subsection{Comparison with state of the art}

In this section, we  compare the performance of Segmenter with respect to the
state-of-the-art methods on ADE20K, Pascal Context and Cityscapes datasets.

\noindent
\textbf{ADE20K.} Seg-B\tsdagger/16 pre-trained on ImageNet-1k  matches the
\textit{state-of-the-art} FCN method DeepLabv3+ ResNeSt-200 \cite{resnest} as shown in Table~\ref{table:ade20k}. Adding our mask transformer, Seg-B\tsdagger-Mask/16 improves by 2\% and achieves a 50.08\% mIoU, outperforming all FCN methods. Our best model, Seg-L-Mask/16 attains a \text{state-of-the-art} performance of 53.63\%, outperforming by a margin of 5.27\% mIoU DeepLabv3+ ResNeSt-200 and the transformer-based methods SETR \cite{setr20} and Swin-L UperNet \cite{swin21}.

\noindent
\textbf{Pascal Context} Table \ref{table:pcontext} reports the performance on
Pascal Context. Seg-B\tsdagger models are competitive with FCN methods and the larger Seg-L/16 model already provides \textit{state-of-the-art} performance, outperforming SETR-L. Performances can be further enhanced with our mask transformer, Seg-L-Mask/16, improving over the linear decoder by 2.5\% and achieving a performance of 59.04\% mIoU. In particular, we report an improvement of 2.8\% over OCR HRNetV2-W48 and 3.2\% over SETR-L MLA.

\noindent
\textbf{Cityscapes.} Table \ref{table:cityscapes_val} reports the performance of
Segmenter on Cityscapes. We use a variant of mask transformer for Seg-L-Mask/16 with only one layer in the decoder as two layers did not fit into memory due to the
large input resolution of $768\!\times\!768$. Both Seg-B and Seg-L methods are competitive with
other \textit{state-of-the-art methods} with Seg-L-Mask/16 achieving a mIoU of 81.3\%.


\noindent
\textbf{Qualitative results.}
Figure \ref{fig:qualitative} shows a qualitative comparison of Segmenter and
DeepLabv3+ with ResNeSt backbone, for which models were provided by the MMSegmentation \cite{mmseg20} library.
We can observe that Deeplabv3+ tends to generate sharper object
boundaries while Segmenter provides more consistent labels on large instances and
handles partial occlusions better. 

\begin{table}
\small
\setlength\tabcolsep{2.7pt} 
\centering
\begin{tabular}{llc}
\toprule
Method & Backbone & mIoU (MS) \\
\midrule
DeepLabv3+  \cite{deeplabv3plus18} & ResNet-101 & 48.5 \\
DANet \cite{fu19dual} & ResNet-101 & 52.6 \\
ANN \cite{zhen19} & ResNet101 & 52.8 \\
CPNet \cite{yu20cpn} & ResNet-101 & 53.9 \\
CFNet \cite{zhang19} & ResNet-101 & 54.0 \\
ACNet \cite{fu19adaptive} & ResNet-101 & 54.1 \\
APCNet \cite{he19} & ResNet101 & 54.7 \\
DNL \cite{yin20} & HRNetV2-W48 & 55.3 \\
DRANet \cite{fu20} & ResNet-101 & 55.4 \\
OCR \cite{yuan20} & HRNetV2-W48 & 56.2 \\
\midrule
SETR-L MLA \cite{setr20} & ViT-L/16 & 55.8 \\
\midrule
\rowcolor{LightCyan}Seg-B\tsdagger/16 & DeiT-B/16 & 53.9 \\
\rowcolor{LightCyan}Seg-B\tsdagger-Mask/16 & DeiT-B/16 & 55.0 \\
\rowcolor{LightCyan}Seg-L/16 & ViT-L/16 & 56.5 \\
\rowcolor{LightCyan}Seg-L-Mask/16 & ViT-L/16 & \textbf{59.0} \\
\bottomrule
\end{tabular}
\vspace{-0.2cm}
\caption{State-of-the-art comparison on Pascal Context validation set.}
\label{table:pcontext}
\vspace{-0.4cm}
\end{table}

\begin{table}
\small
\setlength\tabcolsep{2.7pt} 
\centering
\begin{tabular}{llc}
\toprule
Method & Backbone & mIoU (MS) \\
\midrule
PSANet \cite{zhao18} & ResNet-101 & 79.1 \\
DeepLabv3+ \cite{deeplabv3plus18} & Xception-71 & 79.6 \\
ANN \cite{zhen19} & ResNet-101 & 79.9 \\
MDEQ \cite{bai20} & MDEQ & 80.3 \\
DeepLabv3+ \cite{deeplabv3plus18} & ResNeSt-101 & 80.4 \\
DNL \cite{yin20} & ResNet-101 & 80.5 \\
CCNet \cite{huang19} & ResNet-101 & 81.3 \\
Panoptic-Deeplab \cite{cheng20} & Xception-71 & 81.5 \\
DeepLabv3+ \cite{deeplabv3plus18} & ResNeSt-200 & \textbf{82.7} \\
\midrule
SETR-L PUP \cite{setr20} & ViT-L/16 & \textbf{82.2} \\
\midrule
\rowcolor{LightCyan}Seg-B\tsdagger/16 & DeiT-B/16 & 80.5 \\
\rowcolor{LightCyan}Seg-B\tsdagger-Mask/16 & DeiT-B/16 & 80.6 \\
\rowcolor{LightCyan}Seg-L/16 & ViT-L/16 & 80.7 \\
\rowcolor{LightCyan}Seg-L-Mask/16 & ViT-L/16 & 81.3 \\
\bottomrule
\end{tabular}
\vspace{-0.2cm}
\caption{State-of-the-art comparison on Cityscapes validation set.}
\label{table:cityscapes_val}
\vspace{-0.6cm}
\end{table}

\section{Conclusion}

This paper introduces a pure transformer  approach for semantic
segmentation. The encoding part builds up on the recent Vision Transformer
(ViT), but differs in that we rely on the encoding of all images patches.
 We observe that the transformer captures the global context very well.  
Applying a simple point-wise linear decoder to the patch encodings already achieves
excellent results.
Decoding with a mask transformer further improves the performance. We believe
that our end-to-end encoder-decoder transformer is a first step towards
a unified approach for semantic segmentation, instance segmentation
and panoptic segmentation.

\section{Acknowledgements}
We thank Andreas Steiner for providing the ViT-Base model trained on $8\times 8$ patches and Gauthier Izacard for the helpful discussions. This work was partially supported by the HPC resources from GENCI-IDRIS  (Grant 2020-AD011011163R1), the Louis Vuitton ENS Chair on Artificial Intelligence, and the French government under management of Agence Nationale de la Recherche as part of the ”Investissements d’avenir” program, reference ANR-19-P3IA-0001 (PRAIRIE 3IA Institute).


{
  \small
  \bibliographystyle{ieee_fullname}
  \bibliography{egbib}
}

\clearpage
\begin{appendices}

This appendix presents additional results.  We study the impact of
ImageNet pretraining on the performance and demonstrate its importance in
Section~\ref{sec:pretraining}.
To gain more insight about our approach Segmenter,
we analyze its attention maps and the learned class embeddings in
Section~\ref{sec:attention}. Finally, we give an additional  qualitative comparison
of Segmenter to DeepLabv3+ on ADE20K, Cityscapes and Pascal Context in
Section~\ref{sec:qual}.

\section{ImageNet pre-training}
\label{sec:pretraining}
To study the impact of ImageNet pre-training on Segmenter, we compare our model pre-trained on
ImageNet with equivalent models trained from scratch. To train from scratch, the weights of
the model are initialized randomly with a truncated normal distribution. We use
a base learning rate of $10^{-3}$ and two training procedures. First, we follow
the fine-tuning procedure and use SGD optimizer with ''poly'' scheduler. Second,
we follow a more standard procedure when training a transformer from scratch
where we use AdamW with a cosine scheduler and a linear warmup for $16K$
iterations corresponding to 10\% of the total number of iterations. Table
\ref{table:ablation_pretraining} reports results for Seg-S/16.  We
observe that when pre-trained on ImageNet-21k using SGD, Seg-S/16 reaches 45.37\%
yielding a 32.9\% improvement over the best randomly initialized model.\\

\begin{table}[h]
\small
\setlength\tabcolsep{2.7pt}
\centering
\begin{tabular}{cccc}
  \toprule
  Method & Pre-training  & Optimizer & mIoU (SS) \\
  \midrule
  Seg-S/16 & None & AdamW  & 4.42\\
  Seg-S/16 & None & SGD & 12.51\\
  Seg-S/16 & ImageNet-21k & AdamW  & 34.77\\
  Seg-S/16 & ImageNet-21k & SGD & \textbf{45.37}\\
\bottomrule
\end{tabular}
\caption{Impact of pretraining on the performance  on  ADE20K validation set.}
\label{table:ablation_pretraining}
\vspace{-0.4cm}
\end{table}

\section{Attention maps and class embeddings}
\label{sec:attention}

\begin{figure*}[t]
  \centering
  \includegraphics[width=\linewidth]{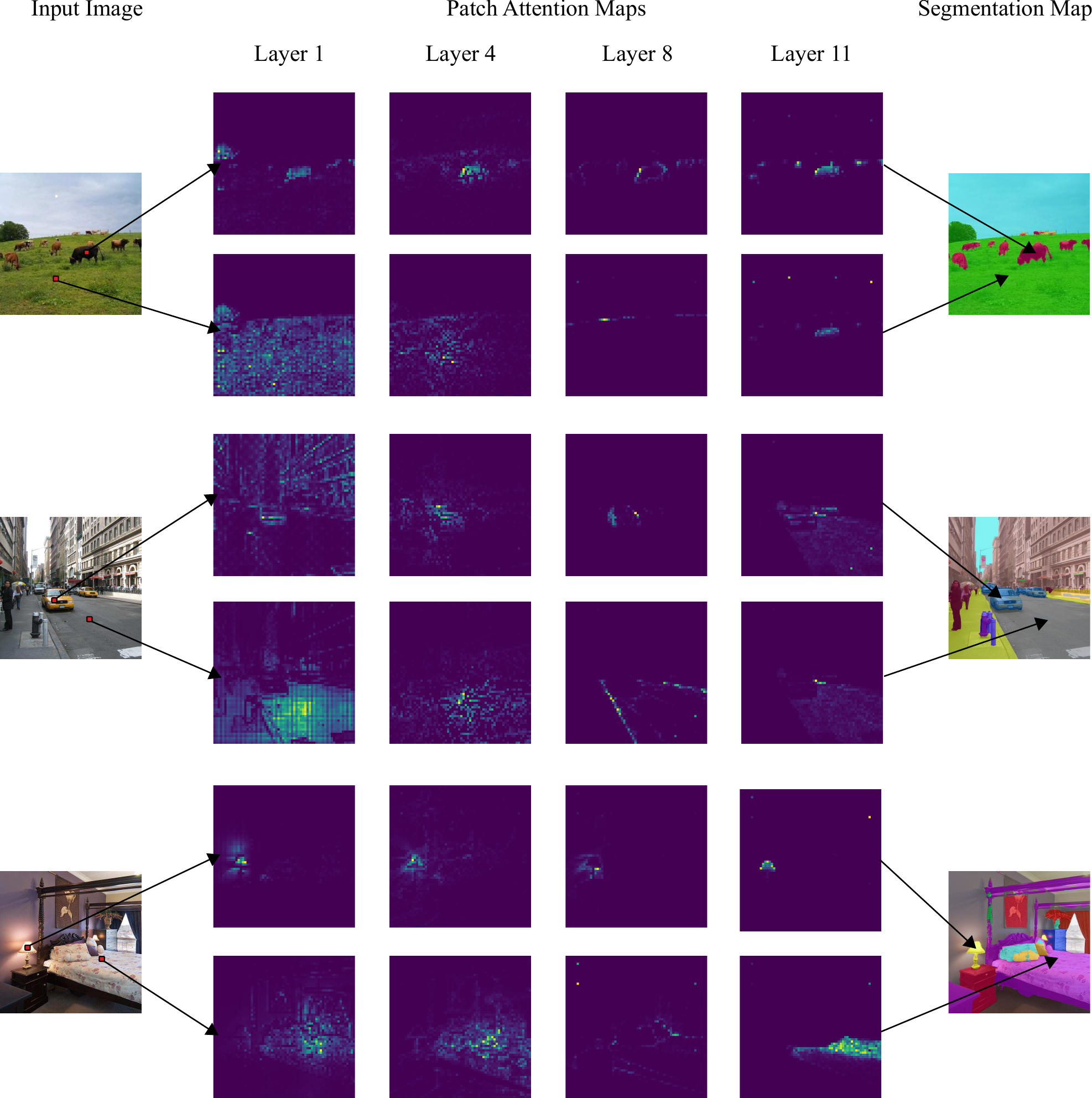}
  \caption{Seg-B/8 patch attention maps for the layers 1, 4, 8 and 11.}
  \label{fig:attention_maps}
\end{figure*}

To better understand how our approach Segmenter processes images, we display
attention maps of Seg-B/8 for 3 images in Figure
\ref{fig:attention_maps}.  We resize attention maps to the original image size. For each image, we analyze attention maps of a patch on a
small instance, for example  lamp, cow or car. We also analyze attention maps of
a patch on a large instance, for example bed, grass and road. We observe that
the attention map field-of-view adapts to the input image and the instance size,
gathering global information on large instances and focusing on local
information on smaller instances. This adaptability is typically not possible
with CNN which have a constant field-of-view, independently of the data.  We
also note there is progressive gathering of information from bottom to top
layers, as for example on the cow instance, where the model first identifies the
cow the patch belongs to, then identifies other cow instances. We observe that attention maps
of lower layers depends strongly on the selected patch while they tend to be more similar for higher layers.

\begin{figure*}[t]
\vspace{-0.4cm}
  \centering
  \includegraphics[width=0.95\linewidth]{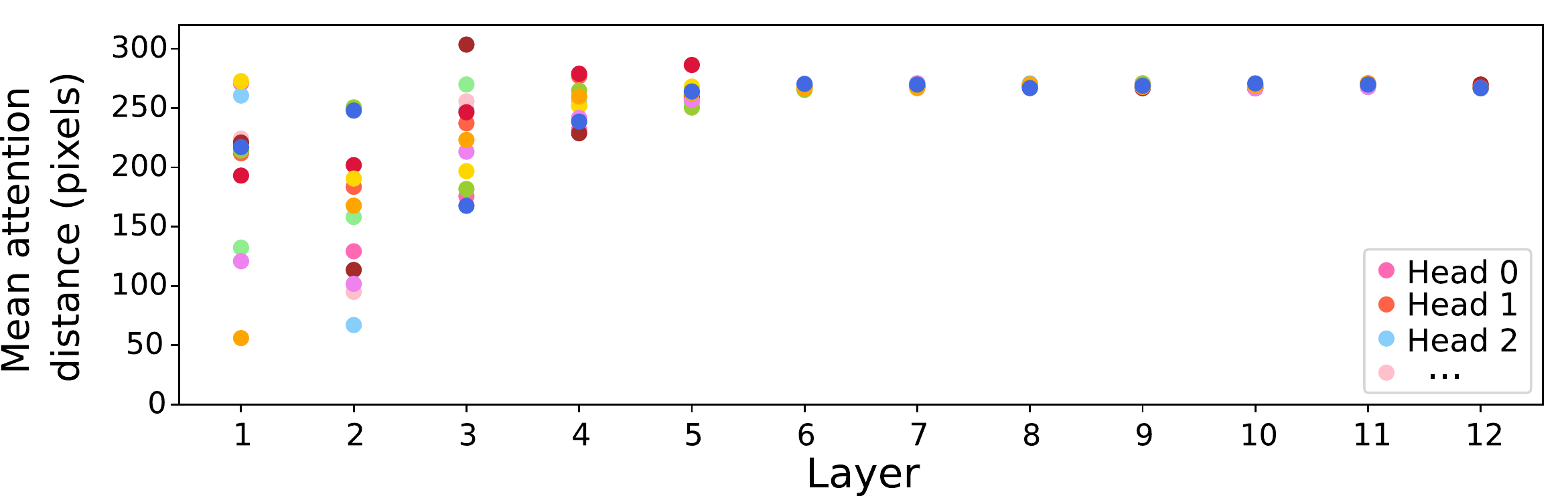}
  \caption{Size of attended area by head and model depth.}
  \label{fig:attention_distance}
  \vspace{-0.4cm}
\end{figure*}

Additionally, to illustrate the larger receptive field size of Segmenter compared to CNNs, we reported the size of the attended
area in Figure \ref{fig:attention_distance}, where each dot shows the mean attention distance for one of the 12 attention heads at each
layer. Already for the first layer, some heads attend to distant patches which clearly lie outside the receptive field of
ResNet/ResNeSt initial layers.

\begin{figure*}[t]
  \centering
  \includegraphics[width=0.8\linewidth]{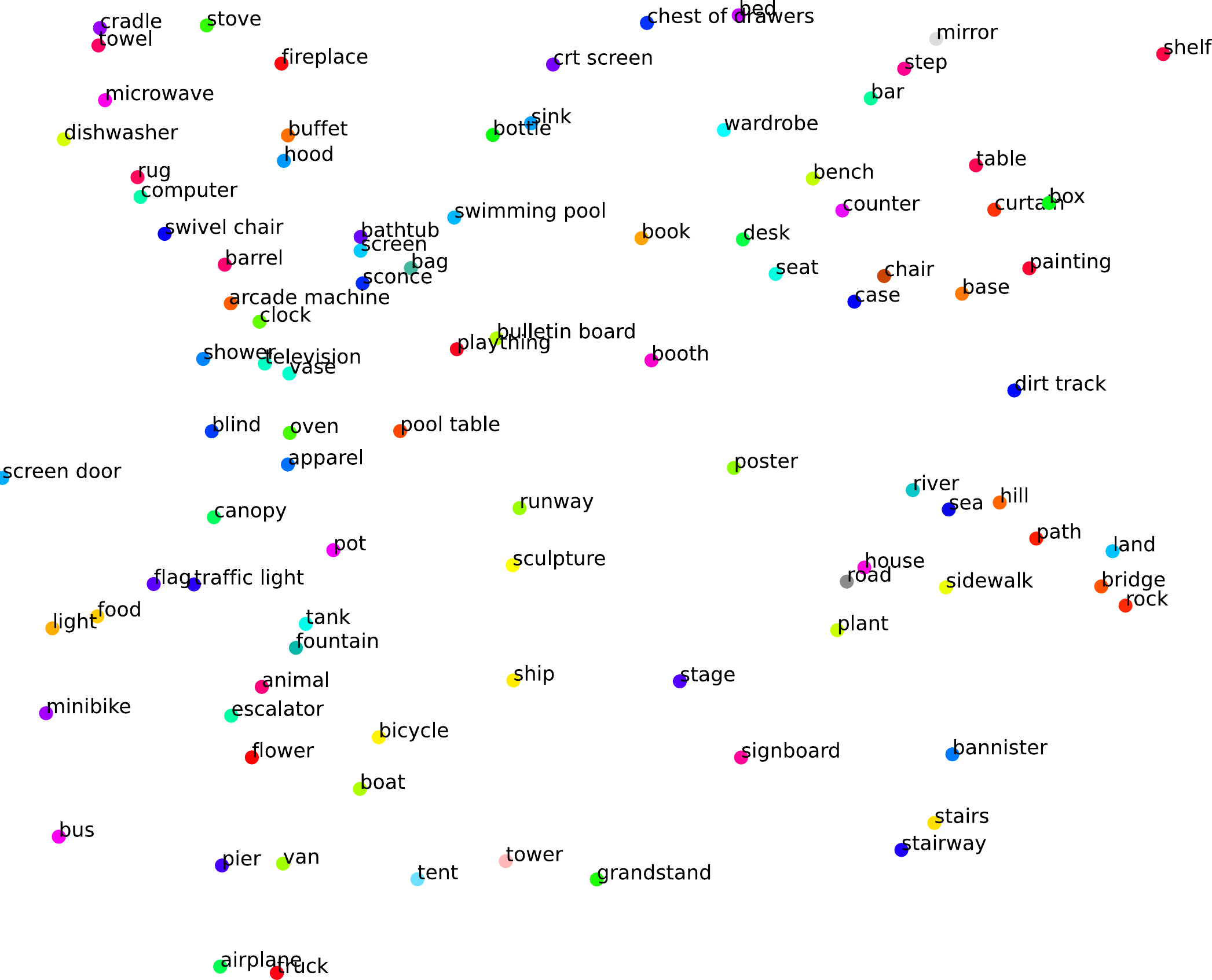}
  \caption{Singular value decompostion of the class embeddings learned with the
    mask transformer on ADE20K.}
  \label{fig:svd}
  \vspace{-0.4cm}
\end{figure*}

To gain some understanding of the class embeddings learned with the mask
transformer, we project embeddings into 2D with a singular value decomposition.
Figure \ref{fig:svd} shows that these projections group instances such as means
of transportation~(bottom left), objects in a house~(top) and outdoor
categories~(middle right). It displays an implicit clustering of semantically
related categories.

\section{Qualitative results}
\label{sec:qual}
\begin{figure*}[t]
  \centering
  \includegraphics[width=0.9\linewidth]{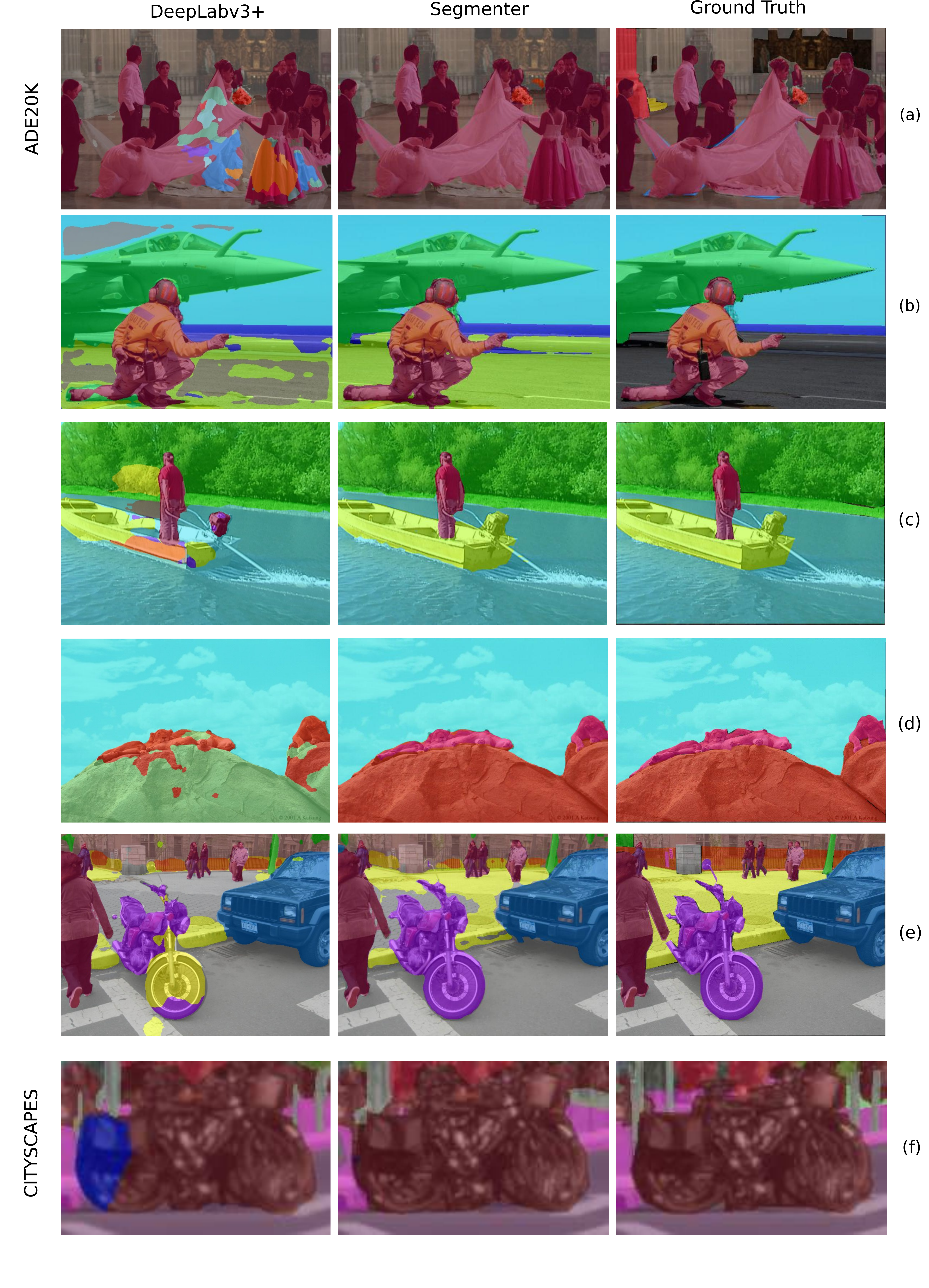}
  \caption{Segmentation maps where Seg-L-Mask/16 produces more coherent segmentation maps than DeepLabv3+ ResNeSt-101.} 
  \label{fig:qualitative_good}
\end{figure*}

\begin{figure*}[t]
  \centering
  \includegraphics[width=\linewidth]{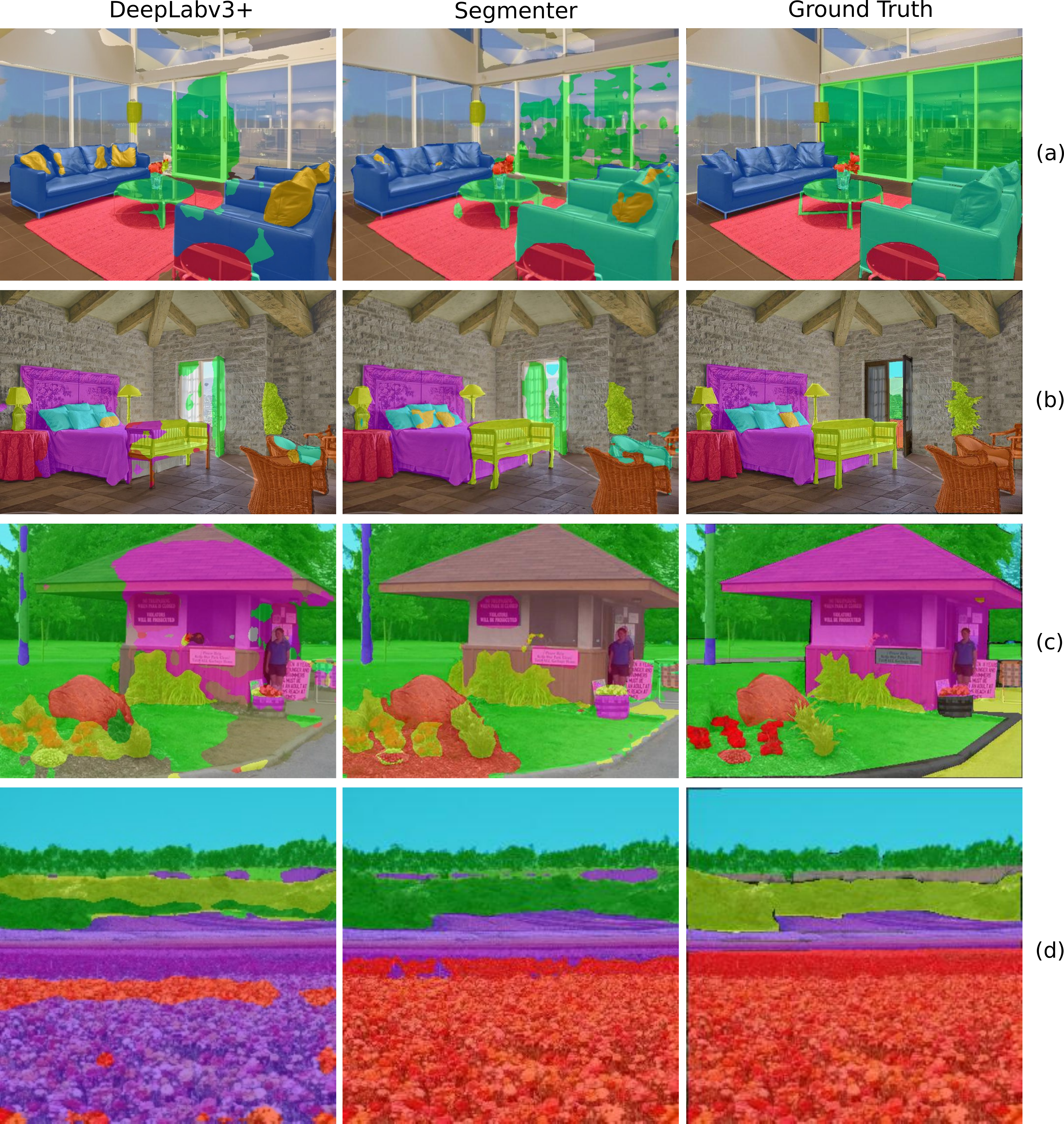}
  \caption{Examples for Seg-L-Mask/16 and DeepLabv3+ ResNeSt-101 on ADE20K, where elements which look very similar are confused.}
  \label{fig:qualitative_mixed_up}
\end{figure*}

\begin{figure*}[t]
  \centering
  \includegraphics[width=\linewidth]{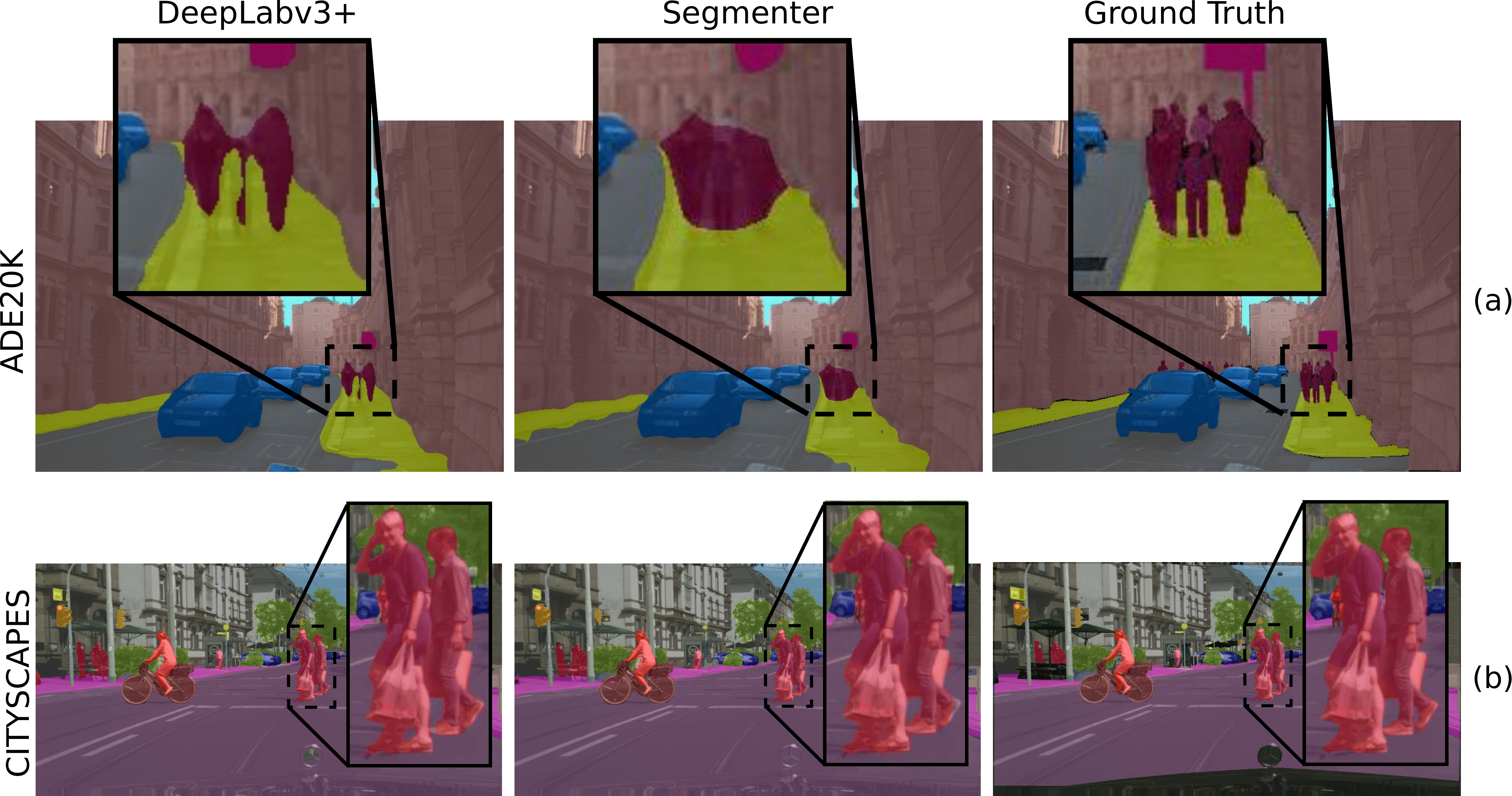}
  \caption{Comparison of Seg-L-Mask/16 with DeepLabV3+ ResNeSt-101 for images with near-by persons. We can observe that DeepLabV3+ localizes boundaries better.}
  \label{fig:qualitative_fine_details}
\end{figure*}

\begin{figure*}[t]
  \centering
  \includegraphics[width=\linewidth]{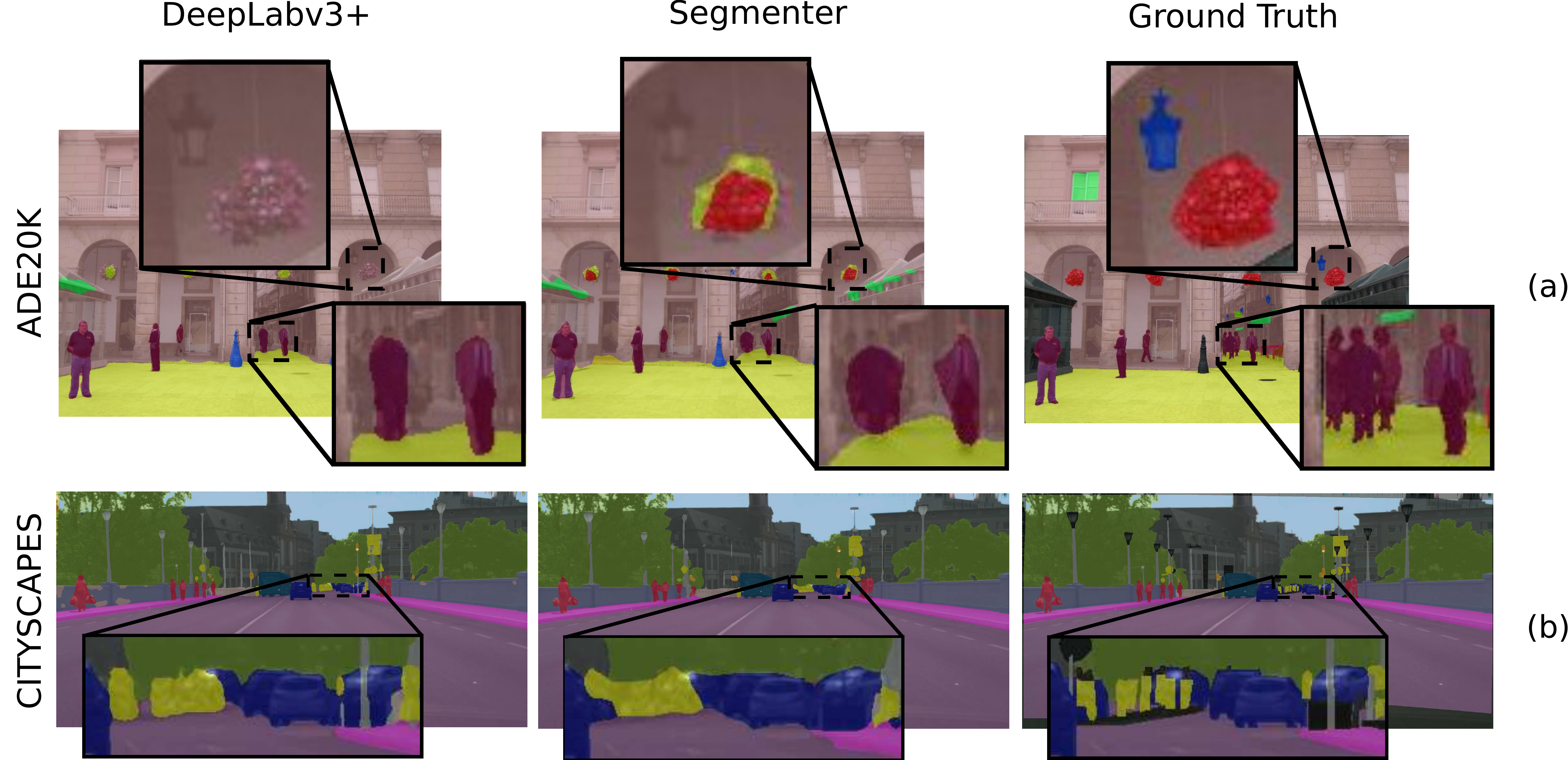}
  \caption{Failure cases of DeepLabV3+ ResNeSt-101 and Seg-L-Mask/16, for small instances such as (a) lamp, people, flowers  and (b)~cars, signals.}
  \label{fig:small_instances}
\end{figure*}

We present additional qualitative results including comparison with DeepLabv3+ ResNeSt-101
and failure cases in Figures \ref{fig:qualitative_good}, \ref{fig:qualitative_mixed_up} and
\ref{fig:qualitative_fine_details}. We can see in Figure \ref{fig:qualitative_good} that
Segmenter produces more coherent segmentation maps than DeepLabv3+. This is the
case for the wedding dress in (a) or the airplane signalmen's helmet in (b). 
In Figure \ref{fig:qualitative_mixed_up}, we show how for some examples, 
different segments which look very similar are confused both in DeepLabv3+ and Segmenter. For example, the armchairs and couchs in  (a), the cushions and pillows in (b) or the trees, flowers and plants in (c) and (d).
In Figure \ref{fig:qualitative_fine_details}, we can see how DeepLabv3+ handles better the
boundaries between different people entities.   Finally,
both Segmenter and DeepLabv3+ have problems segmenting small instances such as
lamp, people or flowers in Figure \ref{fig:small_instances} (a) or the cars and
signals in Figure \ref{fig:small_instances} (b).

\end{appendices}

\end{document}